\documentclass{article}

\usepackage[english]{babel}

\usepackage[letterpaper,top=2cm,bottom=2cm,left=3cm,right=3cm,marginparwidth=1.75cm]{geometry}

\usepackage{amsmath}
\usepackage{amsfonts}
\usepackage{graphicx}
\usepackage[colorlinks=true, allcolors=blue]{hyperref}
\usepackage{caption}

\usepackage{authblk}

\usepackage{shortcuts}

\usepackage[square]{natbib}

\title{MultiView Independent Component Analysis with Delays}

\author[1]{Ambroise Heurtebise}
\author[2]{Pierre Ablin}
\author[1]{Alexandre Gramfort}
\affil[1]{Université Paris-Saclay, Inria, CEA, Palaiseau, France}
\affil[2]{Apple, Paris, France}

\date{}

\begin{document}
\maketitle

\begin{abstract}
Linear Independent Component Analysis (ICA) is a blind source separation technique that has been used in various domains to identify independent latent sources from observed signals. In order to obtain a higher signal-to-noise ratio, the presence of multiple views of the same sources can be used. In this work, we present MultiView Independent Component Analysis with Delays (MVICAD). This algorithm builds on the MultiView ICA model by allowing sources to be delayed versions of some shared sources: sources are shared across views up to some unknown latencies that are view- and source-specific. Using simulations, we demonstrate that MVICAD leads to better unmixing of the sources. Moreover, as ICA is often used in neuroscience, we show that latencies are age-related when applied to Cam-CAN, a large-scale magnetoencephalography (MEG) dataset. These results demonstrate that the MVICAD model can reveal rich effects on neural signals without human supervision.
\end{abstract}

\section{Introduction}\label{sec:introduction}

Independent Component Analysis (ICA) allows to separate mixed signals without knowing the mixing operator~\citep{comon1994independent, hyvarinen2000independent}. In linear ICA~\citep{hyvarinen1999fast, ablin2018faster}, the mixing is assumed linear and the goal is to find a linear transformation of the mixed signals that maximizes the statistical independence of the resulting sources. The parameters of the transformation matrix are estimated by optimizing a cost function, such as a likelihood.

In real-world data, one can sometimes access multiple views from the same data. For example, a doctor may have an MRI scan, a CT scan, and the answers to a clinical questionnaire for a diseased patient. In the context of ICA, a naive method would be to use ICA on each view and compare the resulting sources, but this solution would not take advantage of the group structure. Thus, many multi-view ICA methods have been developed to take advantage of multiple data views and increase the signal-to-noise ratio of the obtained sources (see Section~\ref{sec:related_works}). As multiple views are used to obtain a better recovery of the sources, it is typically assumed that the sources are the same for all views. 

MultiView Independent Component Analysis (MVICA) \citep{richard2020modeling} is an algorithm that makes the latter assumption. It aims to jointly estimate the transformation matrices for multiple views while assuming that they all share the exact same sources. While MVICA guarantees the identifiability of its model and produces state-of-the-art results for several machine learning tasks, it assumes that the sources are perfectly identical and temporally aligned between views. In other words, MVICA ignores the view-specific variability of the sources, which can lead to suboptimal separation results for certain applications.

This kind of inter-view variability typically arises in neuroscience, a critical application of ICA. Indeed, ICA has been extensively used for the analysis of electroencephalography (EEG)~\citep{makeig1995independent} and magnetoencephalography (MEG) signals~\citep{vigario2000independent}, which are non-invasive recordings of the electrical potentials and magnetic fields produced by neurons in the brain. Since these signals are a mixture of the brain's complex neural activity, ICA has been successfully used to separate out the different sources of activity. In this context, using data from multiple subjects at once can reveal better insights about brain functional organization. Yet, neural sources can present temporal delays between different subjects due to individual differences in brain anatomy and functional connectivity~\citep{price2017age, roberts2010meg}. These delays can affect the performance of standard multi-view ICA algorithms, as they assume that the sources are uncorrelated and have the exact same temporal structure across subjects.

In this work, building on the MVICA model, our goal is to develop an ICA algorithm that is robust to temporal delays.  Specifically, we propose a novel method, called MultiView Independent Component Analysis with Delays (MVICAD), that incorporates an iterative temporal alignment step in the estimation procedure. We demonstrate the effectiveness of our method using simulations and MEG data, showing that it outperforms the MVICA algorithm. Besides, we report that the estimated delays of certain sources are related to age.

\section{Related works}\label{sec:related_works}

Many ICA algorithms have been proposed to process data from multiple datasets at once. They vary by their assumptions: some methods assume that the datasets have the same dimension, while others consider that all datasets share the same sources. In this section, we describe some of these algorithms.

MVICA~\citep{richard2020modeling} is our reference algorithm. It is designed to identify shared independent components across multiple datasets. It assumes that the sources in each dataset are independent and that the shared sources have the same temporal patterns across datasets. Unlike some other multi-view ICA approaches, MVICA's likelihood can be written in closed form and is then optimized with a quasi-Newton algorithm, using the gradient and an approximation of the Hessian of the likelihood. Since comparisons with other multi-view ICA methods have been
made in~\citep{richard2020modeling}, we compare our method to MVICA.

Group ICA refers to a whole category of ICA algorithms for processing multiple views of the same data. It usually consists in first applying Principal Component Analysis (PCA) on each view separately. Then all the resulting sets are assembled into one set, using either concatenation or multi-set CCA~\citep{kettenring1971canonical, correa2010canonical}, and PCA is used again on the obtained set. Finally, an ICA algorithm is used. Concatenation can be either spatial or temporal: temporal concatenation produces individual sources and a common mixing matrix, whereas spatial concatenation gives common sources and individual mixing matrices. Since we look for shared sources, we can compare our method to Group ICA with spatial concatenation~\citep{calhoun2009review}. Adding a back-reconstruction step is also possible to recover individual sources. However, this method does not bring statistical guarantees like consistency or asymptotic efficiency, contrary to MVICA. It also gives poorer results than MVICA on several neuroimaging tasks~\citep{richard2020modeling}. One strength of CCA-based ICA~\citep{varoquaux2009canica, tsatsishvili2015combining} is that they project multiple datasets onto a common underlying space by maximizing correlations between sets, which can be used for data fusion. Some of these methods can handle datasets with different numbers of sources. However, taking temporal delays into account in Group ICA would only be possible in the assembling step (done with PCA or multi-set CCA), which is less explicit than doing it during ICA optimization as we do in this paper. 

Shared and Individual Multiview Independent Component Analysis~\citep{pandeva2022multi} can identify both shared and individual sources at the same time. This method assumes that the sources are independent within each dataset but allows for both shared and individual sources to contribute to the data. It uses a joint estimation approach to simultaneously estimate the shared and individual components and their corresponding spatial and temporal maps. However, the number of shared components needs to be specified, which can be difficult.

\section{METHOD}
\label{sec:method}

\textbf{Notation} The absolute value of the determinant of a matrix $W$ is $\vert W \vert$. The $\ell_2$ norm of a vector $\mathbf{s}$ is $\|\mathbf{s}\|$, and $\|S\|$ is the Frobenius norm of a matrix $S$. For a scalar valued function $f$ and a matrix $S \in \bbR^{p \times n}$, we write $f(S) = \sum_{j=1}^p \sum_{i=1}^n f(S_{ji})$.

\subsection{Model and likelihood}
\label{ssec:model}

Given $m$ views and $p$ sources, we model the observed signals $X^i \in \bbR^{p \times n}$, $i=1, \dots, m$, as a linear combination of shared but delayed sources plus noise. Here $n$ is the number of observations (assumed to be common to each view) and $p$ is the number of features, typically corresponding to the number of channels for multivariate signals. The model is:
\begin{equation}\label{eq:model}
    X^i = A^i (\cT_{\btau^i}(S) + N^i) \enspace, \quad i=1, \dots, m \enspace,
\end{equation}
where the $A^i \in \bbR^{p \times p}$ are view-specific mixing matrices, assumed to be full-rank, $\cT$ is a shift operator that delays sources across samples, $S \in \bbR^{p \times n}$ are the shared sources, the vector $\btau^i \in \{-\btau_{\max}, \dots, \btau_{\max}\}^p$ contains the integer delays for the $p$ sources for a given view and $N^i \in \bbR^{p \times n}$ are the noise matrices. In practice, if there are $2$ sources and delays are $\{1, 2\}$, the shift operator will shift to the right the first source by $1$ sample and the second source by $2$ samples. The hyperparameter $\btau_{\max}$ thus represents the maximum shift allowed. These delays represent a temporal shift rather than a time distortion, so they correspond to what is called ``constant delays'' and not to ``cumulative delays'' in \citep{roberts2010meg}.

\begin{figure}[t]
    \centering
    \centerline{\includegraphics[width=\linewidth]{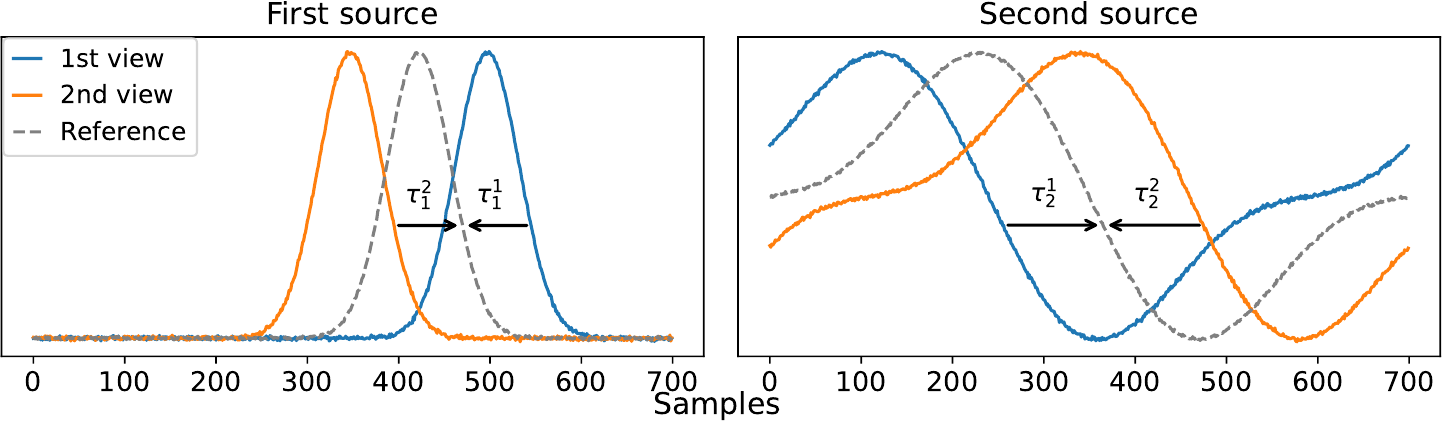}}
    \caption{Sources of 2 views. We observe different delays $\tau^i_j$ for each view $i$ and each source $j$. Blue curves represent the first view's sources and orange ones represent the sources of the second view. Left: first source. Right: second source.}
    \label{fig:sources_2_subjects}
\end{figure}

Figure~\ref{fig:sources_2_subjects} aims to help understand how delays are modeled. It shows how delay can differ for every source and every view: in the figure, the first view has a first source that peaks later than the second view but has an earlier peak for the second source. Also, we observe that the parameters $\tau^i_j$, where $i$ is the view and $j$ is the source, model the delay between sources and the average source, represented in a dotted grey line. 

As usual in ICA, we assume that observations are i.i.d., and so are the sources. Also, we assume that the noise is Gaussian decorrelated of variance $\sigma^2$ and independent across views and from the sources. In practice, thanks to MVICA's robustness to noise misspecification~\citep{richard2020modeling}, estimating the noise level is not critical, so we usually set it to 1.

Importantly, our model is identifiable (see Appendix~\ref{sec:appendix}) up to scale and permutation for the mixing matrices and up to a common delay for the estimated delays. Therefore recovering the sources is a well-posed problem.

We insist on the fact that we represent signals as matrices containing all time samples instead of representing them as vectors. We made that choice because of the use of the circular shift operator $\cT$. Indeed, here we assume periodic boundary conditions, so this operator takes as inputs $p$ signals of length $n$ and $p$ integer delays and rolls each signal by its corresponding delay. Thus, we prefer to consider all observations at once. Having periodic boundary conditions on real signals can sometimes produce unwanted abrupt discontinuities when signals are noisy at the edges of the time interval. This however can be mitigated by applying a windowing function pushing signals to zero at the edges. In particular, this is appropriate in neuroscience for EEG or MEG signals because sources are supposed to emerge from baseline level and go back to baseline at the end of the period of interest.

We derive a maximum-likelihood approach to recover the parameters of the model, in a very similar way to~\citep{richard2020modeling}. As usual in ICA, sources reconstruction is done by estimating first the unmixing matrices $W^i = (A^i)^{-1} \in \bbR^{p \times p}$. Thus, we see the likelihood as a function of $\mathbf{W} = \{W^1, \dots, W^m\}$ and $\btau = \{\btau^1, \dots, \btau^m\}$. The negative log-likelihood can be written as:
\begin{equation}\label{eq:likelihood}
    \mathcal{L} (\mathbf{W}, \boldsymbol{\btau}) = - \sum_{i=1}^m \log \vert W^i \vert + \frac{1}{2 \sigma^2} \sum_{i=1}^m \| Y^i - \bar{S} \|^2 + f(\bar{S}) ,
\end{equation}
where $Y^i = \cT_{-\btau^i}(W^i X^i)$ are the aligned estimated sources of subject $i$, $\bar{S} = \frac1m \sum_{i=1}^m Y^i$ are the average estimated sources and $f$ is a smoothed version of the logarithm of the source density $d$ by convolution with a Gaussian kernel. As explained in~\citep{jung1997extended}, the density $d$ does not need to be known and can be estimated easily as soon as we know if sources are sub- or super-Gaussian. In neuroscience, we commonly assume that neural sources are super-Gaussian.

\subsection{Unmixing matrix optimization}
\label{ssec:unmixing_optimization}

Optimizing Equation~(\ref{eq:likelihood}) with respect to all parameters at once is complex since the variables $\btau^i$ are discrete, while the $W^i$ are continuous, so we choose to minimize it iteratively by block coordinate descent. First, $\mathcal{L}$ is alternatively minimized with respect to each $W^i$. Subsequently, the minimization process takes place alternately with respect to each $\tau^i$, and so forth.

By applying $\cT_{\btau^i}(\cdot)$ in the squared norm and in function $f$, optimization of $\mathcal{L}$ with respect to $W^i$ boils down to minimizing:
\begin{align}
    \mathcal{L}^i (W^i) 
    &= - n \: \log \vert W^i \vert \nonumber \\ 
    &+ \frac{m-1}{2 m \sigma^2} \left\| W^i X^i - \sum_{j \neq i} \frac{\cT_{\btau^i - \btau^j}(W^j X^j)}{m-1} \right\|^2 \nonumber \\
    &+ f \left( \frac1m W^i X^i + \frac1m \sum_{j \neq i} \cT_{\btau^i - \btau^j}(W^j X^j) \right) . \nonumber
\end{align}

This formula strongly looks like Equation (4) in~\citep{richard2020modeling}. Indeed, the only difference lies in the fact that we obtained $\cT_{\btau^i - \btau^j}(W^j X^j)$ instead of $W^j X^j$ in both sums over\:$j$. Intuitively, in the squared norm, we compare $W^i X^i$ to the quantity $\frac{1}{m-1} \sum_{j \neq i} \cT_{\btau^i - \btau^j}(W^j X^j)$, where the last term is the shared source estimate of all subjects but subject $i$, deliberately delayed by subject $i$'s delays. Having similar formulas allows us to derive the relative gradient expression and Hessian approximation of $\mathcal{L}$ with respect to $W^i$ immediately. To do so, we need to replace the estimated sources $W^j X^j$ of subject $j \neq i$ by its shifted version $\cT_{\btau^i - \btau^j}(W^j X^j)$ in the formulas of gradient $G^i \in \bbR^{p \times p}$ and Hessian approximation $H^i \in \bbR^{p \times p \times p \times p}$ in~\citep{richard2020modeling}. Then, we use a quasi-Newton approach by computing a direction $D = -(H^i)^{-1} G^i$ and finding with line search a step size $\rho$ such that $\cL^i((I_p + \rho D)W^i) < \cL^i(W^i)$.

\subsection{Delay optimization}
\label{ssec:delay_optimization}

Equation~\ref{eq:likelihood} should be optimized with respect to $\btau^i$. Because of its complex form, the term $f(\bar{S})$ is burdensome to minimize rapidly, so we chose to discard it. Indeed, minimization of the true loss function would require an exhaustive search that has a computational complexity of $o(\btau_{\max} \times n)$, while using only the squared norm part will allow us to use cross-correlation to quickly compute the error, hence a computational complexity of $o(\btau_{\max} \times \log(n))$. As the first term, $-\sum_{i=1}^m \log |W^i|$, is independent from $\btau^i$, the only part that needs to be minimized is: 
$$
\mathcal{L}^i{}' (\btau^i) 
= \sum_{i=1}^m \| Y^i - \bar{S} \|^2
= \sum_{i=1}^m \sum_{j=1}^p \| \mathbf{y}^i_j - \bar{\mathbf{s}}_j \|^2 \enspace ,
$$
where $\mathbf{y}^i_j \in \bbR^n$ (resp. $\bar{\mathbf{s}}_j \in \bbR^n$) is the $j$-th row of $Y^i$ (resp. $\bar{S}$). Since $\mathbf{y}^i_j - \bar{\mathbf{s}}_j$ only depends on $\btau^i_j$, and not on $\btau^i_k$ for $k \neq j$, we can minimize $\mathcal{L}^i{}'$ with respect to each $\btau^i_j$ separately.

Thus, optimization boils down to finding, for each source $j \in \{1, \dots, p\}$ separately, the integer $\btau^i_j$ that maximizes $\langle \mathbf{y}^i_j, \mathbf{y}^{-i}_j \rangle$, where $\mathbf{y}^{-i}_j = \frac{1}{m-1} \sum_{k \neq i} \mathbf{y}^k_j$. This step is simply done by cross-correlation. Stacking the results in a vector gives $\btau^i = \{\btau^i_1, \dots, \btau^i_p\}$ and concludes the delay optimization step of subject $i$. 

Note that we have to set a maximum delay parameter $\btau_{\max}$ in MVICAD. This parameter represents the maximum delay possible, in absolute value, for each source and each subject, that our algorithm will test. In the neuroscience application, there are natural choices for $\btau_{\max}$ (see Section~\ref{ssec:real_data}).

Furthermore, we compared on simulations our method with an exhaustive search that includes the last term of Equation~\ref{eq:likelihood} in delay optimization. We observed that ignoring this term does not affect performance. 

\section{EXPERIMENTS}
\label{sec:experiments}

All the code is written in Python and is available on our \href{https://github.com/AmbroiseHeurtebise/mvicad}{GitHub repo}.

\subsection{Simulation study}
\label{ssec:synthetic}

To evaluate the performance of the MVICAD algorithm, we conducted a number of simulations allowing us to quantify the error in parameter estimation.

In synthetic experiments, we simulate data according to our model~(\ref{eq:model}). We use $5$ subjects, $3$ sources, and $700$ samples, as this number of samples is comparable to the one available for MEG data (See below).

\begin{figure}[t]
    \centering
    \centerline{ \includegraphics[width=\linewidth]{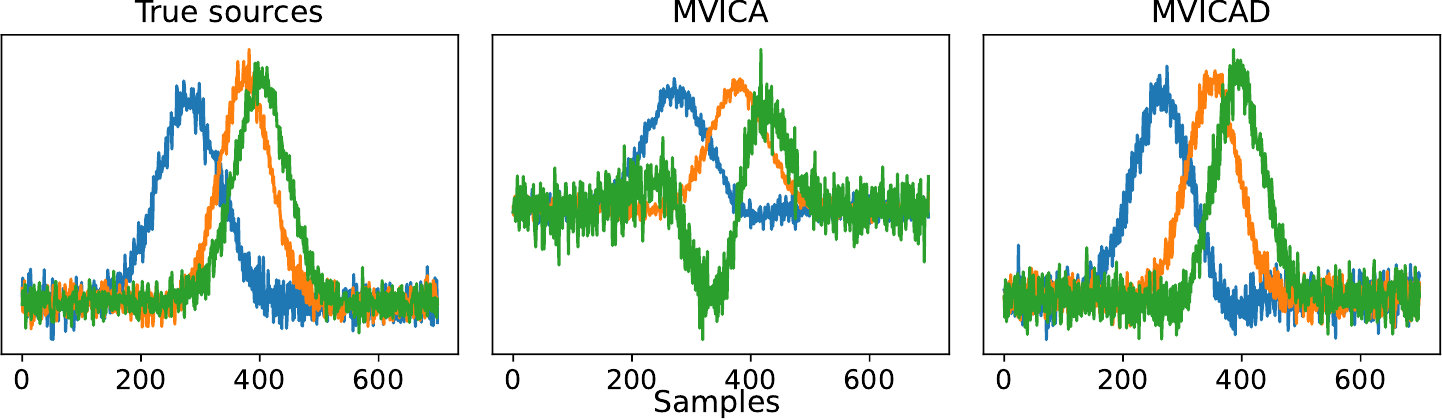}}
    \caption{Left: true generated shared sources. Middle: sources estimated by MVICA. Right: sources estimated by MVICAD.}
    \label{fig:sources}
\end{figure}

In Figure~\ref{fig:sources}, we also use a maximum delay parameter of $40$, as this number corresponds to $40$\:ms if the sampling rate is $1000$\:Hz, which is the typical maximum amount of delay in the context of MEG signals used in real data experiments~\citep{price2017age}.
This figure (left) shows the true simulated shared sources. In the middle, we observe that sources found by MVICA do not fit these true sources, contrary to the sources found by MVICAD (right).

\subsubsection{Amari distance}
\label{sssec:amari}

\begin{figure}[t]
    \centering
    \centerline{\includegraphics[width=8.5cm]{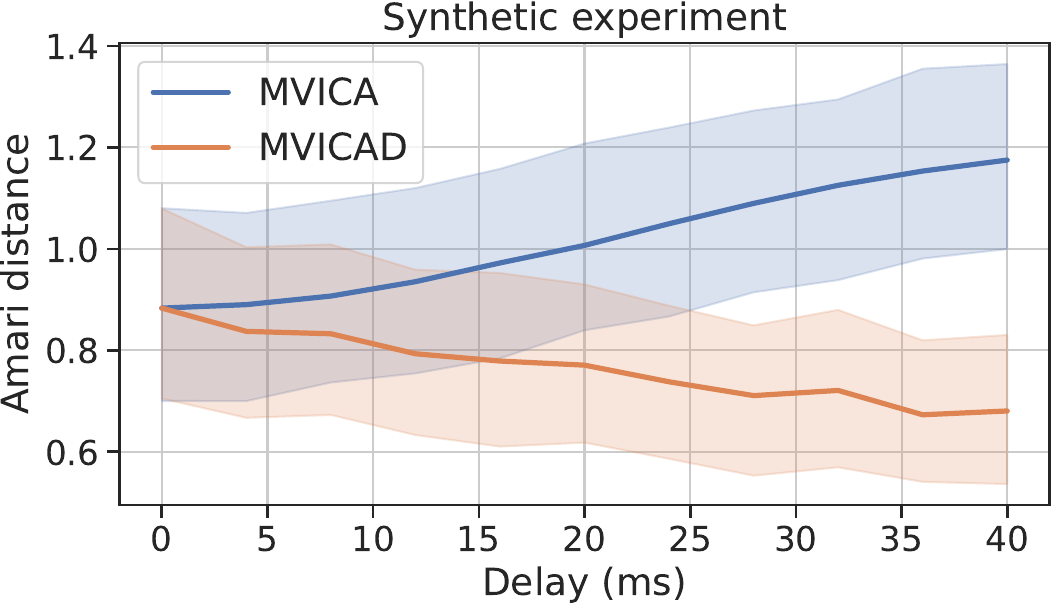}}
    \caption{Amari distance of both MVICA and MVICAD for several levels of delay. We observe that our algorithm outperforms MVICA as soon as some delay is introduced in the model.}
    \label{fig:amari}
\end{figure}

As classically done for ICA algorithms, we use the Amari distance~\citep{moreau1998self} to quantify the error on mixing matrices. The benefit of this metric is that it is scale and permutation invariant.
Figure~\ref{fig:amari} represents the Amari distance of both MVICA and MVICAD for several levels of delay introduced in the model. We see that our algorithm outperforms MVICA in this particular setup, as soon as there is some delay. In fact, the curves seem to be approximately linear. They start from the same point when there is no delay, and when there is a delay of $40$, MVICA's Amari distance is $1.18$, whereas the one of MVICAD is equal to $0.68$.

Note that, for each level of delay in the $x$-axis, the maximum delay parameter of MVICAD is set to this level of delay precisely. Consequently, both algorithms have the same Amari distance when the delay equals $0$. Indeed, when the maximum delay parameter is set to $0$, our algorithm strictly corresponds to MVICA. 

\subsubsection{Artificial delays}
\label{sssec:artificial}

Our next experiment illustrates the ability of the model to recover delays. We used $40$ subjects, $5$ sources, $700$ samples, a maximum delay parameter of $40$, and a quantity of noise that gave a signal-to-noise ratio approximately equal to 5.

Figure~\ref{fig:scatter_plot_art} shows the results of this experiment. In total, there are $40 \times 5 = 200$ estimated delays. We observe that estimated delays are strongly correlated with artificial delays: the slope of the fit function is almost $1$, and so does the $R^2$ score.

Note that not having a slope of exactly $1$ can be partially explained by the presence of the noise. Searching a delay for each source of each subject means having $mp$ delay parameters and is thus sensitive to noise. Also, it can be explained by the fact that the number of samples is rather low ($700$).
Nevertheless, this experiment shows the ability of MVICAD to retrieve the true delays.

\subsection{Real data experiments}
\label{ssec:real_data}

In order to test MVICAD on real data, we use the Cam-CAN dataset. In this dataset delays usually do not exceed $60$\:ms for visual tasks and $20$\:ms for auditory tasks~\citep{price2017age}. Since the sampling rate is equal to $1000$\:Hz in the data, we set $\btau_{\max}$ to $60$ (number of samples).

\subsubsection{Cam-CAN dataset}
\label{sssec:camcan}

The Cam-CAN dataset~\citep{taylor2017cambridge} is a large and comprehensive dataset of neuroimaging, cognitive, and demographic data collected from a group of healthy adults. The dataset includes data from $661$ participants, ranging in age from 18 to 88 years old, with equal numbers of participants in each $10$-year age range. The participants were recruited from the general population in the Cambridgeshire area of the UK, with the aim of recruiting a sample that was representative of the local population in terms of age, sex, and education level.

This dataset is freely available to researchers and has become a resource for investigating the neurobiological mechanisms underlying healthy aging and age-related cognitive decline.

The Cam-CAN Stage 2 repository contains a subset of data from the larger Cam-CAN dataset, focused on the second wave of assessments conducted approximately two years after the initial assessment. The MEG data in the Stage 2 repository includes recordings of brain activity using a whole-head $306$ channel Elekta Neuromag Vectorview system. It contains data from several cognitive tasks, including visual and auditory tasks. For each of these tasks, participants were presented a series of stimuli, and averaging the obtained signals give what is called evoked data. 

In our experiments, we focus on evoked data of both tasks. We had to preprocess data and remove subjects with unusual signals, which reduced the number of subjects for the visual and auditory tasks. Also, these data last $0.7$ second at a sampling rate of $1000$ Hz, thus giving datasets of shape $(477, 306, 701)$ for the visual task and $(501, 306, 701)$ for the auditory one.

\begin{figure}
    \centering
    \begin{minipage}{.46\textwidth}
      \centering
      \includegraphics[width=\linewidth]{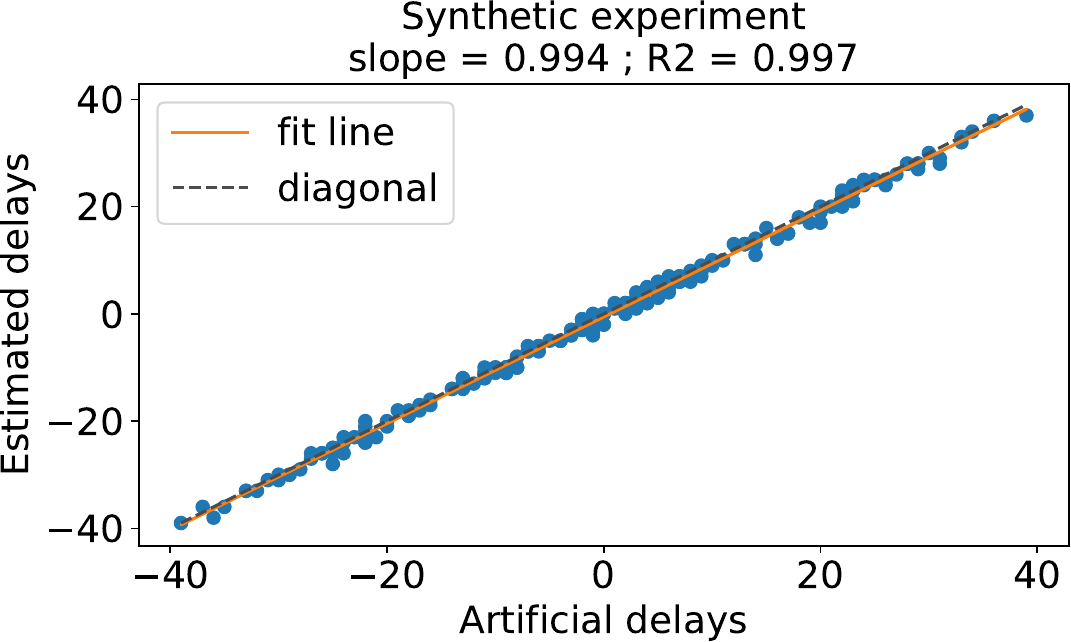}
      \captionof{figure}{Scatter plot of artificial and estimated delays. We used $40$ subjects and $5$ sources, and each point represents one source of one subject.}
      \label{fig:scatter_plot_art}
    \end{minipage}%
    \hfill
    \begin{minipage}{.46\textwidth}
      \centering
      \includegraphics[width=\linewidth]{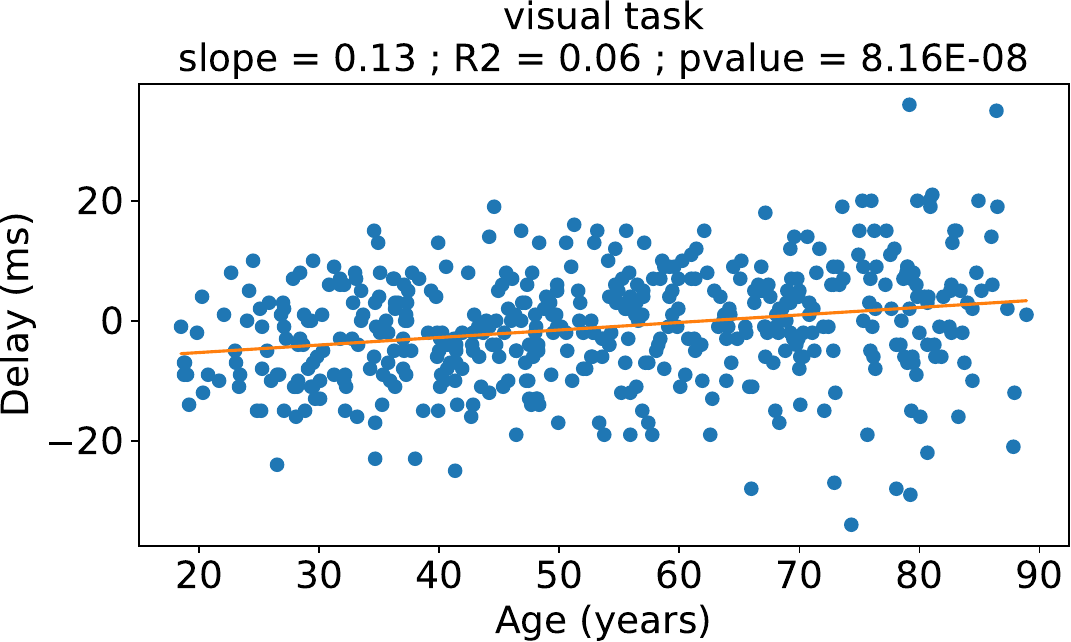}
      \captionof{figure}{Scatter plot of age and delay of the $477$ subjects of the visual dataset. We observe a significant correlation between aging and estimated delay.}
      \label{fig:scatter_plot_age}
    \end{minipage}
\end{figure}

\subsubsection{Delay and age correlation}
\label{sssec:scatterplot}

The next experiment consists in studying age-delay correlation, as Cam-CAN data are often used to investigate mechanisms underlying healthy aging and age-related cognitive decline.

Figure~\ref{fig:scatter_plot_age} shows the scatter plot of age and delay of the 477 subjects of the visual task. This scatter plot shows a significant correlation between the age of subjects and estimated delays. To produce it, we first applied a PCA with $10$ components on the dataset, whose shape became $(477, 10, 701)$, and then used MVICAD to get estimated delays of shape $(477, 10)$. We report here the source whose delays are most linearly related to age (using $R^2$ score, and p-value of Pearson correlation). From a biological point of view, having relevant and non-relevant sources in terms of age-delay correlation makes sense because some of the $10$ obtained sources represent noise, whereas others are closer to the actual neural signals affected by aging.

However, we can also investigate the model with only one delay per subject. In other words, we force estimated delays to be equal across sources in our algorithm, i.e. we only estimate one delay per subject. This method reveals some aging effects without having to visually explore delays of each source, yet it is not source-specific. The obtained results are: slope $ = 0.08$ ; $R^2 = 0.08$ ; p-value $ = 9.69 \times 10^{-11}$. The drawback of having one delay per subject is that it produces a higher negative log-likelihood and poorer source reconstruction.

Finally, note that this experiment produces lower results for the auditory task. It was expected since auditive stimuli are known to present ``cumulative delays'' instead of ``constant delays'' \citep{roberts2010meg}.

\section{CONCLUSION}
\label{sec:conclusion}

We have proposed an improvement of the algorithm MVICA which aims to retrieve latent sources from multiple views and can be used in various domains. To do so, we estimated delays that appear in each source of each subject and use them to correct our source reconstruction during optimization. We presented the likelihood of our model in closed form and proposed a way of minimizing it with respect to both unmixing matrices and delays. In the context of MEG signals, we showed that our estimated delays were significantly correlated to age and demonstrated on a synthetic experiment that our method outperforms MVICA in terms of Amari distance.

\section{ACKNOWLEDGEMENT}
\label{sec:acknowledgement}

Data collection and sharing for this project was provided by the Cambridge Centre for Ageing and Neuroscience (CamCAN). CamCAN funding was provided by the UK Biotechnology and Biological Sciences Research Council (grant number BB/H008217/1), together with support from the UK Medical Research Council and University of Cambridge, UK. Data used in the preparation of this work were obtained from the CamCAN repository (available at \url{http://www.mrc-cbu.cam.ac.uk/datasets/camcan/}).
This work was supported by the ANR BrAIN (ANR-20-CHIA0016) grant.

\newpage

\bibliographystyle{apalike}
\bibliography{refs}

\begin{thebibliography}{}

\bibitem[Ablin et~al., 2018]{ablin2018faster}
Ablin, P., Cardoso, J.-F., and Gramfort, A. (2018).
\newblock Faster independent component analysis by preconditioning with hessian approximations.
\newblock {\em IEEE Transactions on Signal Processing}, 66(15):4040--4049.

\bibitem[Calhoun et~al., 2009]{calhoun2009review}
Calhoun, V.~D., Liu, J., and Adal{\i}, T. (2009).
\newblock A review of group {ICA} for {fMRI} data and {ICA} for joint inference of imaging, genetic, and {ERP} data.
\newblock {\em Neuroimage}, 45(1):S163--S172.

\bibitem[Comon, 1994]{comon1994independent}
Comon, P. (1994).
\newblock Independent component analysis, a new concept?
\newblock {\em Signal processing}, 36(3):287--314.

\bibitem[Correa et~al., 2010]{correa2010canonical}
Correa, N.~M., Adali, T., Li, Y.-O., and Calhoun, V.~D. (2010).
\newblock Canonical correlation analysis for data fusion and group inferences.
\newblock {\em IEEE signal processing magazine}, 27(4):39--50.

\bibitem[Hyvarinen, 1999]{hyvarinen1999fast}
Hyvarinen, A. (1999).
\newblock Fast and robust fixed-point algorithms for independent component analysis.
\newblock {\em IEEE transactions on Neural Networks}, 10(3):626--634.

\bibitem[Hyv{\"a}rinen and Oja, 2000]{hyvarinen2000independent}
Hyv{\"a}rinen, A. and Oja, E. (2000).
\newblock Independent component analysis: algorithms and applications.
\newblock {\em Neural networks}, 13(4-5):411--430.

\bibitem[Jung et~al., 1997]{jung1997extended}
Jung, T.-P., Humphries, C., Lee, T.-W., Makeig, S., McKeown, M., Iragui, V., and Sejnowski, T.~J. (1997).
\newblock Extended {ICA} removes artifacts from electroencephalographic recordings.
\newblock {\em Advances in neural information processing systems}, 10.

\bibitem[Kettenring, 1971]{kettenring1971canonical}
Kettenring, J.~R. (1971).
\newblock Canonical analysis of several sets of variables.
\newblock {\em Biometrika}, 58(3):433--451.

\bibitem[Makeig et~al., 1995]{makeig1995independent}
Makeig, S., Bell, A., Jung, T.-P., and Sejnowski, T.~J. (1995).
\newblock Independent component analysis of electroencephalographic data.
\newblock {\em Advances in neural information processing systems}, 8.

\bibitem[Moreau and Macchi, 1998]{moreau1998self}
Moreau, E. and Macchi, O. (1998).
\newblock Self-adaptive source separation. ii. comparison of the direct, feedback, and mixed linear network.
\newblock {\em IEEE transactions on signal processing}, 46(1):39--50.

\bibitem[Pandeva and Forr{\'e}, 2022]{pandeva2022multi}
Pandeva, T. and Forr{\'e}, P. (2022).
\newblock Multi-view independent component analysis with shared and individual sources.
\newblock {\em arXiv preprint arXiv:2210.02083}.

\bibitem[Price et~al., 2017]{price2017age}
Price, D., Tyler, L.~K., Neto~Henriques, R., Campbell, K.~L., Williams, N., Treder, M.~S., Taylor, J.~R., and Henson, R. (2017).
\newblock Age-related delay in visual and auditory evoked responses is mediated by white-and grey-matter differences.
\newblock {\em Nature communications}, 8(1):15671.

\bibitem[Richard et~al., 2020]{richard2020modeling}
Richard, H., Gresele, L., Hyvarinen, A., Thirion, B., Gramfort, A., and Ablin, P. (2020).
\newblock Modeling shared responses in neuroimaging studies through multiview {ICA}.
\newblock {\em Advances in Neural Information Processing Systems}, 33:19149--19162.

\bibitem[Roberts et~al., 2010]{roberts2010meg}
Roberts, T.~P., Khan, S.~Y., Rey, M., Monroe, J.~F., Cannon, K., Blaskey, L., Woldoff, S., Qasmieh, S., Gandal, M., Schmidt, G.~L., et~al. (2010).
\newblock {MEG} detection of delayed auditory evoked responses in autism spectrum disorders: towards an imaging biomarker for autism.
\newblock {\em Autism Research}, 3(1):8--18.

\bibitem[Taylor et~al., 2017]{taylor2017cambridge}
Taylor, J.~R., Williams, N., Cusack, R., Auer, T., Shafto, M.~A., Dixon, M., Tyler, L.~K., Henson, R.~N., et~al. (2017).
\newblock {The Cambridge Centre for Ageing and Neuroscience (Cam-CAN) data repository: Structural and functional MRI, MEG, and cognitive data from a cross-sectional adult lifespan sample}.
\newblock {\em neuroimage}, 144:262--269.

\bibitem[Tsatsishvili et~al., 2015]{tsatsishvili2015combining}
Tsatsishvili, V., Cong, F., Toiviainen, P., and Ristaniemi, T. (2015).
\newblock Combining {PCA} and multiset {CCA} for dimension reduction when group {ICA} is applied to decompose naturalistic {fMRI} data.
\newblock In {\em 2015 International Joint Conference on Neural Networks (IJCNN)}, pages 1--6. IEEE.

\bibitem[Varoquaux et~al., 2009]{varoquaux2009canica}
Varoquaux, G., Sadaghiani, S., Poline, J.~B., and Thirion, B. (2009).
\newblock {CanICA}: Model-based extraction of reproducible group-level {ICA} patterns from {fMRI} time series.
\newblock {\em arXiv preprint arXiv:0911.4650}.

\bibitem[Vig{\'a}rio et~al., 2000]{vigario2000independent}
Vig{\'a}rio, R., Sarela, J., Jousmiki, V., Hamalainen, M., and Oja, E. (2000).
\newblock Independent component approach to the analysis of {EEG} and {MEG} recordings.
\newblock {\em IEEE transactions on biomedical engineering}, 47(5):589--593.

\end{thebibliography}

\newpage

\appendix

\section{Appendix}
\label{sec:appendix}

Let us prove the identifiability of our model. We consider multiple i.i.d. r.v. $\bs_k \in \mathbb{R}^p$, $k=1, \dots, \btau_{\max}$, that model sources, instead of only one r.v. $\bs$, as is often the case. Having multiple variables allows us to delay them with operator $\cT$. For simplicity, we define $\bs := \bs_1$. Recall that our model is:
$$
\bx^i = A^i (\cT_{\btau^i}(\bs) + \bn^i) \enspace, \quad i=1, \dots, m \enspace,
$$
with $\btau^i \in \mathbb{R}^p$. Note that, compared to Equation (\ref{eq:model}), we used vectors instead of matrices. Assume that we also have:
$$
\bx^i = A'^i (\cT_{\btau'^i}(\bs') + \bn'^i) \enspace, \quad i=1, \dots, m \enspace,
$$
for some $A'^i, \btau'^i, \bs'$ and $\bn'^i$. Note that we also define $\bs' := \bs'_1$ where $\bs'_k \in \mathbb{R}^p$, $k=1, \dots, \btau_{\max}$, are i.i.d. r.v.

Since we assume that $\bs_k$ has non-Gaussian independent components (whose densities are not reduced to a point-like mass) and that the r.v. $\bs_k$ are i.i.d. and that $\bn^i$ is Gaussian decorrelated, then $\cT_{\btau^i}(\bs) + \bn^i$ also has non-Gaussian independent components. And so does $\cT_{\btau'^i}(\bs') + \bn'^i$. Following \citep{comon1994independent}, Theorem $11$, there exists a scale-permutation matrix $P^i$ such that $A'^i = A^i P^i$. As a consequence, and since $A^i$ is invertible, we have, for all~$i$:
\begin{align}
    \cT_{\btau^i}(\bs) + \bn^i 
    & = P^i (\cT_{\btau'^i}(\bs') + \bn'^i) \nonumber \\
    \Rightarrow \bs + \cT_{-\btau^i}(\bn^i) 
    & = \cT_{-\btau^i} \left( P^i (\cT_{\btau'^i}(\bs') + \bn'^i) \right) \nonumber \\
    & = P^i (\cT_{\btau'^i - (P^i)^\top \btau^i}(\bs') - \cT_{-(P^i)^\top \btau^i}(\bn'^i)). \nonumber
\end{align}
The last line comes from the equality $\cT_{\btau} P = P \cT_{P^\top \btau}$ for a scale-permutation matrix $P$ and a vector $\btau$.

We focus on subject $1$ and subject $i \neq 1$:
\begin{align}
    & \bs + \cT_{-\btau^1}(\bn^1) - (\bs + \cT_{-\btau^i}(\bn^i)) \nonumber \\
    & = P^1 (\cT_{\btau'^1 - (P^1)^\top \btau^1}(\bs') + \cT_{-(P^1)^\top \btau^1}(\bn'^1)) \nonumber \\
    & - P^i (\cT_{\btau'^i - (P^i)^\top \btau^i}(\bs') + \cT_{-(P^i)^\top \btau^i}(\bn'^i)). \nonumber
\end{align}
Consequently,
\begin{align}
    & P^1 \cT_{\btau'^1 - (P^1)^\top \btau^1}(\bs') - P^i \cT_{\btau'^i - (P^i)^\top \btau^i}(\bs') \nonumber \\
    & = \cT_{-\btau^1}(\bn^1) - \cT_{-\btau^i}(\bn^i) \nonumber \\
    & + P^i \cT_{-(P^i)^\top \btau^i}(\bn'^i) - P^1 \cT_{-(P^1)^\top \btau^1}(\bn'^1). \label{eq:right_gauss}
\end{align}
Since the right-hand side of the last equality is a linear combination of Gaussian random variables, this implies that $P^1 \cT_{\btau'^1 - (P^1)^\top \btau^1}(\bs') - P^i \cT_{\btau'^i - (P^i)^\top \btau^i}(\bs')$ is also Gaussian.

Let us show that there exists a vector $\btau \in \mathbb{R}^p$ such that, for all~$i$, $\btau'^1 - (P^1)^\top \btau^1 = \btau'^i - (P^i)^\top \btau^i =: \btau$. Note that this is an equality modulo $\btau_{\max}$. Indeed, delaying sources by $\btau$ and by $\btau + q \btau_{\max}$, $q \in \mathbb{Z}$, is equivalent, according to our definition of the shift operator $\cT$. For simplicity, we omit the modulo part in equalities about delays. 

By contradiction, let us suppose that $\btau'^1 - (P^1)^\top \btau^1 \neq \btau'^i - (P^i)^\top \btau^i$. Let us call $C \subset \{1, \dots, p\}$ the set of indices such that $(\btau'^1 - (P^1)^\top \btau^1)_C = (\btau'^i - (P^i)^\top \btau^i)_C$ and let $\bar{C} = \{1, \dots, p\} \backslash C$. By assumption, $\bar{C} \neq \emptyset$. By definition of $C$, $\left( \cT_{\btau'^1 - (P^1)^\top \btau^1}(\bs') \right)_C = \left( \cT_{\btau'^i - (P^i)^\top \btau^i}(\bs') \right)_C$, and $\left( \cT_{\btau'^1 - (P^1)^\top \btau^1}(\bs') \right)_{\bar{C}}$ and $\left( \cT_{\btau'^i - (P^i)^\top \btau^i}(\bs') \right)_{\bar{C}}$ are independent. 

Thus, the left-hand side of Equation (\ref{eq:right_gauss}) only depends on three terms: $\left( \cT_{\btau'^1 - (P^1)^\top \btau^1}(\bs') \right)_C$,\linebreak$\left( \cT_{\btau'^1 - (P^1)^\top \btau^1}(\bs') \right)_{\bar{C}}$ and $\left( \cT_{\btau'^i - (P^i)^\top \btau^i}(\bs') \right)_{\bar{C}}$, which are all three independent from each others.

The left-hand side of Equation (\ref{eq:right_gauss}) is thus a linear combination of independent random variables and is Gaussian. So, by Cramér's lemma, we should have that $\cT_{\btau'^1 - (P^1)^\top \btau^1}(\bs')$ and $\cT_{\btau'^i - (P^i)^\top \btau^i}(\bs')$ are Gaussian too, which is absurd, given that $\bs'$ is assumed to be non-Gaussian. So, there exists a vector $\btau$ such that $\btau'^1 - (P^1)^\top \btau^1 = \btau'^i - (P^i)^\top \btau^i =: \btau$.

Let us define $\mathbf{\tilde{s}} := \cT_{\btau'^1 - (P^1)^\top \btau^1}(\bs') = \cT_{\btau'^i - (P^i)^\top \btau^i}(\bs')$. Since the right-hand side of Equation (\ref{eq:right_gauss}) is Gaussian, it follows that $(P^1 - P^i) \: \mathbf{\tilde{s}}$ is Gaussian too. Thus, the equality holds only if $P^1 = P^i$. Therefore, the matrices $P^i$ are all equal, and there exists a scale-permutation matrix $P$ such that $A'^i = A^i P$.

In conclusion, we proved that there exists a scale and permutation matrix $P$ and a vector $\btau \in \mathbb{R}^p$ such that, for all $i$, $A'^i = A^i P$ and $\btau'^i = P^\top \btau^i + \btau$.

\end{document}